\documentclass{llncs}
\usepackage{amsmath,amssymb,calc,ifthen}
\usepackage{float}
\usepackage[table,usenames,dvipsnames]{xcolor} 
\usepackage{tikz}
\usepackage{hyperref}
\usepackage{url}
\hypersetup{
colorlinks,
citecolor=black,
filecolor=black,
linkcolor=black,
urlcolor=black
}
\usetikzlibrary{plotmarks,shapes}
\usepackage{amsmath,graphicx}
\usepackage{epstopdf}
\usepackage{subcaption}
\usepackage{graphicx}
\usepackage{color}

\usepackage{listings}
\usepackage{pdfpages}
\usepackage{enumitem} 

\usepackage[linesnumbered,noend]{algorithm2e}

\SetCommentSty{mycommfont}

\DeclareMathOperator*{\argmin}{arg\,min}

\begin{document}

\definecolor{blue3}{HTML}{86B7FC} 
\definecolor{blue1}{HTML}{B5F1FF} 
\definecolor{blue2}{HTML}{E0F9FF} 

\title{Disease Knowledge Transfer across Neurodegenerative Diseases}
\titlerunning{Disease Knowledge Transfer across Neurodegenerative Diseases}  
%

\author{R\u{a}zvan V. Marinescu\inst{1,2} \and Marco Lorenzi\inst{5} \and Stefano B. Blumberg\inst{1} \and Alexandra L. Young\inst{1} \and Pere Planell-Morell\inst{1} \and Neil P. Oxtoby\inst{1} \and Arman Eshaghi\inst{1,3} \and Keir X. Yong\inst{4} \and Sebastian J. Crutch\inst{4} \and Polina Golland\inst{2} \and Daniel C. Alexander\inst{1}, for the Alzheimer's Disease Neuroimaging Initiative} 


\authorrunning{} 


\institute{Centre for Medical Image Computing, University College London, UK
\and 
Computer Science and Artificial Intelligence Laboratory, MIT, USA
\and
Queen Square MS Centre, UCL Institute of Neurology, UK
\and 
Dementia Research Centre, University College London, UK
\and
University of C\^{o}te d'Azur, Inria Sophia Antipolis, France
}

\maketitle              

\newcommand{\expFld}{figures}

\begin{abstract}
We introduce Disease Knowledge Transfer (DKT), a novel technique for transferring biomarker information between related neurodegenerative diseases. DKT infers robust multimodal biomarker trajectories in rare neurodegenerative diseases even when only limited, unimodal data is available, by transferring information from larger multimodal datasets from common neurodegenerative diseases. DKT is a joint-disease generative model of biomarker progressions, which exploits biomarker relationships that are shared across diseases. Our proposed method allows, for the first time, the estimation of plausible \emph{multimodal} biomarker trajectories in Posterior Cortical Atrophy (PCA), a rare neurodegenerative disease where only unimodal MRI data is available. For this we train DKT on a combined dataset containing subjects with two distinct diseases and sizes of data available: 1) a larger, multimodal typical AD (tAD) dataset from the TADPOLE Challenge, and 2) a smaller unimodal Posterior Cortical Atrophy (PCA) dataset from the Dementia Research Centre (DRC), for which only a limited number of Magnetic Resonance Imaging (MRI) scans are available. Although validation is challenging due to lack of data in PCA, we validate DKT on synthetic data and two patient datasets (TADPOLE and PCA cohorts), showing it can estimate the ground truth parameters in the simulation and predict unseen biomarkers on the two patient datasets. While we demonstrated DKT on Alzheimer's variants, we note DKT is generalisable to other forms of related neurodegenerative diseases. Source code for DKT is available online: \url{https://github.com/mrazvan22/dkt}.

\end{abstract}

\section{Introduction}

The estimation of accurate biomarker signatures in Alzheimer's disease (AD) and related neurodegenerative diseases is crucial for understanding underlying disease mechanisms, predicting subjects' progressions, and enrichment in clinical trials. Recently, data-driven disease progression models were proposed to reconstruct long term biomarker signatures from collections of short term individual measurements \cite{oxtoby2018,jedynak2012computational}. When applied to large datasets of typical AD, disease progression models have shown important benefits in understanding the earliest events in the AD cascade \cite{oxtoby2018}, quantifying biomarkers' heterogeneity \cite{young2018uncovering} and they showed improved predictions over standard approaches \cite{oxtoby2018}. However, by necessity these models require large datasets -- in addition they should be both multimodal and longitudinal. Such data is not always available in rare neurodegenerative diseases. In particular, most datasets for rare neurodegenerative diseases come from local clinical centres, are unimodal (e.g. MRI only) and limited both cross-sectionally and longitudinally -- this makes the application of disease progression models extremely difficult.  Moreover, such a model estimated from common diseases such as typical AD may not generalise to specific variants. For example, in Posterior Cortical Atrophy (PCA) -- a neurodegenerative syndrome causing visual disruption -- posterior regions such as the occipital lobe are affected early, instead of the hippocampus and temporal regions in typical AD.

The problem of limited data in medical imaging has so far been addressed through transfer learning methods. These were successfully used to improve the accuracy of AD diagnosis \cite{hon2017towards} or prediction of MCI conversion \cite{cheng2015domain}, but have two key limitations. First, they use deep learning or other machine learning methods, which are not easily interpretable and don't allow us to understand underlying disease mechanisms that are either specific to rare diseases, or shared across related diseases. Secondly, these models cannot be used to forecast the future evolution of subjects at risk of disease, which is important for selecting the right subjects in clinical trials. 

We propose Disease Knowledge Transfer (DKT), a generative model that estimates continuous multimodal biomarker progressions for multiple diseases simultaneously -- including rare neurodegenerative diseases -- and which inherently performs transfer learning between the modelled phenotypes. This is achieved by exploiting biomarker relationships that are shared across diseases, whilst accounting for differences in the spatial distribution of brain pathology. DKT is interpretable, which allows us to understand underlying disease mechanisms, and can also predict the future evolution of subjects at risk of diseases. We apply DKT on Alzheimer's variants and demonstrate its ability to predict non-MRI trajectories for patients with Posterior Cortical Atrophy, in lack of such data. This is done by fitting DKT to two datasets simultaneously: (1) the TADPOLE Challenge \cite{marinescu2018tadpole} dataset containing subjects from the Alzheimer's Disease Neuroimaging Initiative (ADNI) with MRI, FDG-PET, DTI, AV45 and AV1451 scans and (2) MRI scans from patients with Posterior Cortical Atrophy from the Dementia Research Centre (DRC), UK. We finally validate DKT on three datasets: 1) simulated data with known ground truth, 2) TADPOLE sub-populations with different progressions and 3) 20 DTI scans from controls and PCA patients from our clinical center.

\begin{figure}[h]
 \centering
 \includegraphics[width=0.9\textwidth,trim=0 0 0 0,clip]{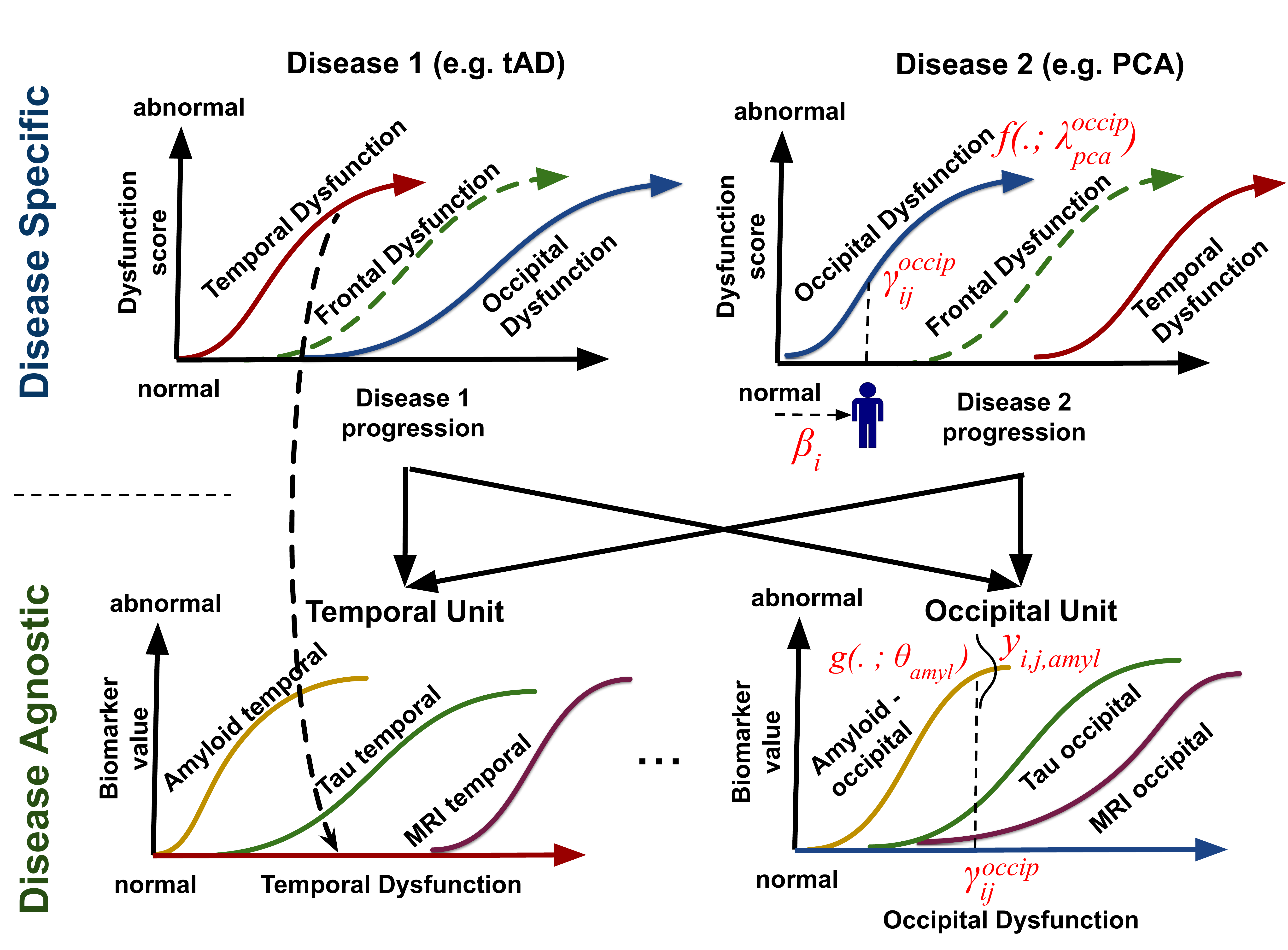}
 \caption{Diagram of the proposed DKT framework. We assume that each disease can be modelled as the evolution of abstract dysfunction scores (Y-axis, top row), each one related to different brain regions. Each region-specific dysfunction score then further models (X-axis, bottom row) the progression of several multimodal biomarkers within that same region. For instance, the temporal dysfunction, modelled as a biomarker in the disease specific model (top row), is the X-axis in the disease agnostic model (temporal unit, bottom row), which aggregates together abnormality from amyloid, tau and MR imaging within the temporal lobe. The biomarker relationships within the bottom units are assumed to be disease agnostic and shared across all diseases modelled. Knowledge transfer between the two diseases can then be achieved via the disease-agnostic units. Mathematical notation from section \ref{sec:method} is shown in red to ease understanding.}
 \label{fig:diagram}
\end{figure}

\section{Method}
\label{sec:method}


\newcommand{\lp}{\lambda_{d_i}^{\psi(k)}}
\newcommand{\lpuu}{\lambda_{d_i}^{\psi(k),(u)}}
\newcommand{\lpum}{\lambda_{d_i}^{\psi(k),(u-1)}}

Fig. \ref{fig:diagram} shows the diagram of the DKT framework. We assume that the progression of each disease can be modelled as a unique evolution of dysfunction trajectories representing region-specific multimodal pathology, further modelled as the progression of several biomarkers within that same region, but acquired using different modalities (Fig. \ref{fig:diagram} bottom). Each group of biomarkers in the bottom row will be called a \emph{disease-agnostic unit} or simply \emph{agnostic unit}, because biomarker dynamics here are assumed to be shared across all diseases modelled.

The assumption that the dynamics of some biomarkers are disease-agnostic (i.e. shared across diseases), is key to DKT. We can make this assumption for two reasons. First, pathology in many related neurodegenerative diseases (e.g. Alzheimer's variants) is hypothesised to share the same underlying mechanisms (e.g. amyloid and tau accumulation), and within one region, such mechanisms lead to similar pathology dynamics across all the disease variants modelled \cite{jack2010hypothetical}, with the key difference that distinct brain regions are affected at different times and with different pathology rates and extent, likely caused by selective vulnerability of networks within these regions \cite{seeley2009neurodegenerative}. Secondly, even if the diseases share different upstream mechanisms (e.g. amyloid vs tau accumulation), downstream biomarkers measuring hypometabolism, white matter degradation and atrophy are likely to follow the same pathological cascade and will have similar dynamics.



We now model the biomarker dynamics that are specific to each disease, by mapping the subjects' disease stages to dysfunction scores. We assume that each subject $i$ at each visit $j$ has an underlying disease stage $s_{ij} = \beta_{i} + m_{ij}$, where $m_{ij}$ represents the months since baseline visit for subject $i$ at visit $j$ and $\beta_{i}$ represents the time shift of subject $i$. We then assume that each subject $i$ at visit $j$ has a dysfunction score $\gamma_{ij}^l$ corresponding to multimodal pathology in brain region $l$, which is a function of its disease stage:

\begin{equation}
\label{eqDysfunctionScoreDef}
 \gamma_{ij}^l = f(\beta_{i} + m_{ij}; \lambda_{d_i}^l)
\end{equation}
where $f$ is a smooth monotonic function mapping each disease stage to a dysfunction score, having parameters $\lambda_{d_i}^l$ corresponding to agnostic unit $l \in \Lambda$, where $\Lambda$ is the set of all agnostic units. Moreover, $d_i \in \mathbb{D}$ represents the index of the disease corresponding to subject $i$, where $\mathbb{D}$ is the set of all diseases modelled. For example, MCI and tAD subjects from ADNI as well as tAD subjects from the DRC cohort can all be assigned $d_i=1$, while PCA subjects can be assigned $d_i=2$.  We implement $f$ as a parametric sigmoidal curve similar to \cite{jedynak2012computational}, to enable a robust optimisation and because this accounts for floor and ceiling effects present in AD biomarkers -- the monotonicity of this sigmoidal family is also very appropriate for many neurodegenerative diseases due to irreversability.


We further model the biomarker dynamics that are disease-agnostic, by constructing the mapping from the dysfunction scores $\gamma_{ij}^l$ to the biomarker measurements. We assume a set of given biomarker measurements $Y = [y_{ijk} | (i,j,k) \in \Omega]$ for subject $i$ at visit $j$ in biomarker $k$, where $\Omega$ is the set of available biomarker measurements. We further denote by $\theta_k$ the trajectory parameters for biomarker $k \in K$ within its agnostic unit $\psi(k)$, where $\psi$: \{1, ..., K\} $ \rightarrow \Lambda$ maps each biomarker $k$ to a unique agnostic unit $l \in \Lambda$. These definitions allow us to formulate the likelihood for a single measurement $y_{ijk}$ as follows:

\begin{equation}
 p(y_{ijk}|\theta_k, \lp, \beta_{i}, \epsilon_k) = N(y_{ijk}| g( \gamma_{ij}^{\psi(k)} ; \theta_k), \epsilon_k)
\end{equation}
where $g(\ .\ ; \theta_k)$ represents the trajectory of biomarker $k$ within agnostic unit $\psi(k)$, with parameters $\theta_k$, and is again implemented using a sigmoidal function for reasons outlined above. Parameters $\lp$ are used to define $\gamma_{ij}^{\psi(k)}$ based on Eq. \ref{eqDysfunctionScoreDef}, where agnostic unit $l$ is now referred to as $\psi(k)$, to clarify this is the unit where biomarker $k$ has been allocated. Variable $\epsilon_k$ denotes the variance of measurements for biomarker $k$.


We extend the above model to multiple subjects, visits and biomarkers to get the full model likelihood:
\begin{equation}
\label{eq:dktFinal}
 p(\boldsymbol{y}|\theta, \lambda, \beta , \epsilon) = \\ \prod_{(i,j,k) \in \Omega} p(y_{ijk}|\theta_k, \lp, \beta_{i}) 
\end{equation}
where $\boldsymbol{y} = [y_{ijk} | \forall (i,j,k) \in \Omega ]$ is the vector of all biomarker measurements, while $\boldsymbol{\theta} = [\theta_1, ..., \theta_K]$ represents the stacked parameters for the trajectories of biomarkers in agnostic units, $\boldsymbol{\lambda} = [\lambda_d^{l}|l \in \Lambda, d \in \mathbb{D}]$ are the parameters of the dysfunction trajectories within the disease models, $\boldsymbol{\beta} =[\beta_1, ..., \beta_N]$ are the subject-specific time shifts and $\boldsymbol{\epsilon} = [\epsilon_k | k \in K]$ estimates measurement noise. 

We estimate the model parameters $[\boldsymbol{\theta}, \boldsymbol{\lambda}, \boldsymbol{\beta}, \boldsymbol{\epsilon}]$ using loopy belief propagation -- see algorithm in supplementary material. One key advantage of DKT is that the subject's time shift $\beta_{i}$ can be estimated using only a subset (e.g. MRI) of the subject's data -- the model can then infer the missing modalities (e.g. non-MRI) using Eq. \ref{eq:dktFinal}.

\subsection{Generating Synthetic Data}
\label{sec:dktMetSyn}

We first test DKT on synthetic data, to assess its performance against known ground truth. More precisely, we generate data that follows the DKT model exactly, and test DKT's ability to recover biomarker trajectories and subject time-shifts. We generate synthetic data from two diseases (50 subjects with "\emph{synthetic PCA}" and 100 subjects with "\emph{synthetic AD}") using the parameters from the bottom-left table in Fig. \ref{fig:dktSynthTrajCompTrue}, emulating the TADPOLE and DRC cohorts -- see supplementary material for full details. The six biomarkers ($k_1$-$k_6$) have been \emph{a-priori} allocated to two agnostic units $l_0$ and $l_1$. To simulate the lack of multimodal data in the synthetic PCA subjects, we discarded the data from biomarkers $k_0$, $k_1$, $k_4$ and $k_5$ for all these subjects. 


\subsection{Data Acquisition and Preprocessing}

We trained DKT on ADNI data from the TADPOLE challenge \cite{marinescu2018tadpole}, since it contained a large number of multimodal biomarkers already pre-processed and aggregated into one table. From the TADPOLE dataset we selected a subset of 230 subjects which had an MRI scan and at least one FDG PET, AV45, AV1451 or DTI scan. In order to model another disease, we further included MRI scans from 76 PCA subjects from the DRC cohort, along with scans from 67 tAD and 87 age-matched controls.
 
For both datasets, we computed multimodal biomarker measurements corresponding to each brain lobe: MRI volumes using the Freesurfer software, FDG-, AV45- and AV1451-PET standardised uptake value ratios (SUVR) extracted with the standard ADNI pipeline, and DTI fractional anisotropy (FA) measures from adjacent white-matter regions. For every lobe, we regressed out the following covariates: age, gender, total intracranial volume (TIV) and dataset (ADNI vs DRC). Finally, biomarkers were normalized to the [0,1] range. 


\section{Results on Synthetic and Patient Datasets}
\label{sec:dktRes}

\begin{figure}[htp]
\includegraphics[width=0.8\textwidth]{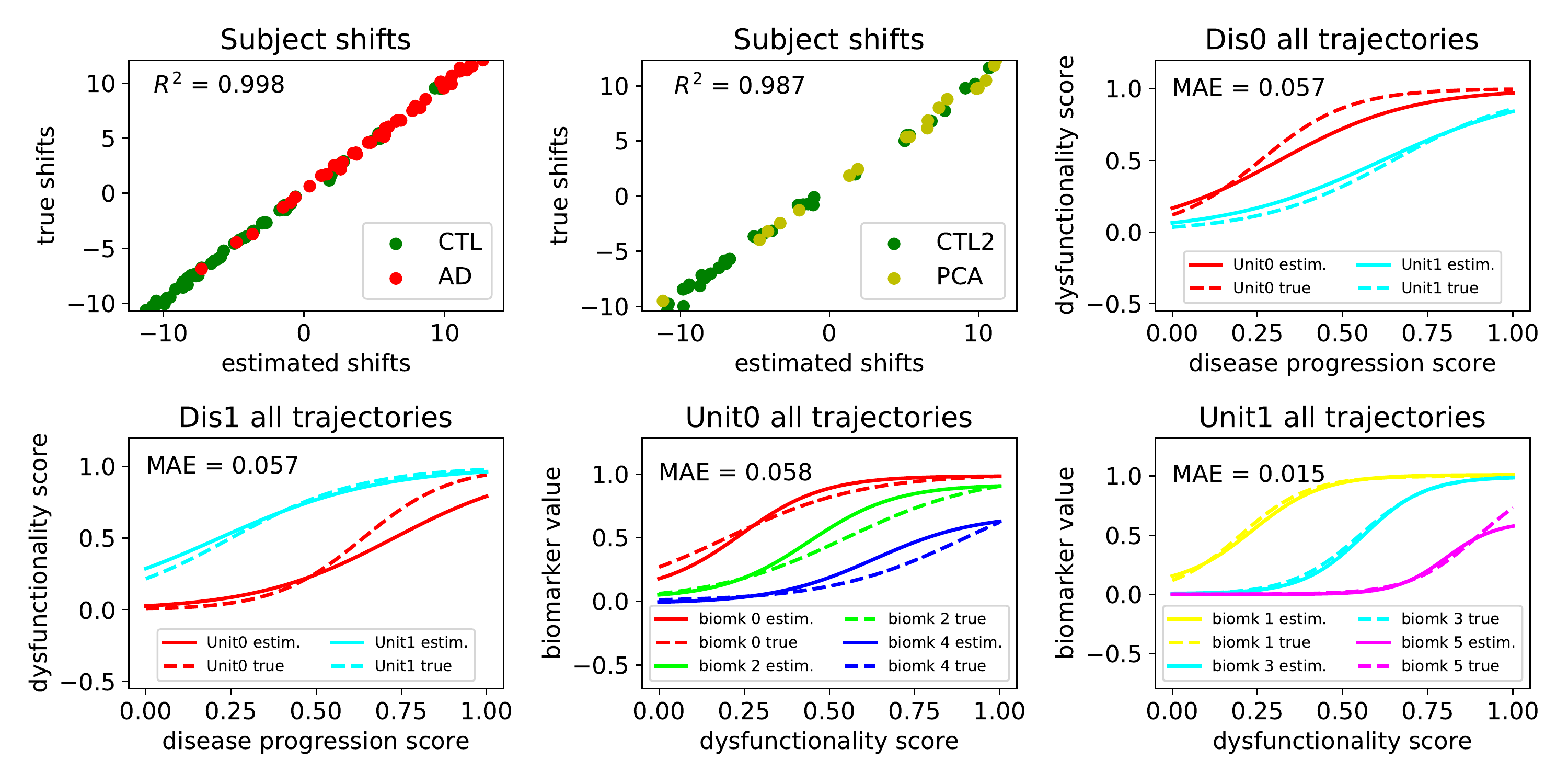}
\begin{flushright}
\vspace{-16.0em}
\fontsize{5}{8.0}\selectfont
 \begin{tabular}{c}
 Biomarker allocation:\\
 $l_0:\{k_0, k_2, k_4\}$ \\
 $l_1: \{k_1, k_3, k_5\}$ \\
\hline
Agnostic unit $l_0$:\\
$\theta_0 = (1,5,0.20,0)$\\
$\theta_2 = (1,5,0.55,0)$\\
$\theta_4 = (1,5,0.90,0)$ \\
\hline
Agnostic unit $l_1$:\\
$\theta_1 = (1,10,0.20,0)$\\
$\theta_3 = (1,10,0.55,0)$\\
$\theta_5 = (1,10,0.90,0)$ \\
\hline  
Synthetic AD:\\
$\lambda_0^0 = (1, 0.3, -4, 0)$\\
$\lambda_0^1 = (1, 0.2,\ \ 6, 0)$ \\
\hline
Synthetic PCA:\\ 
$\lambda_1^0 = (1, 0.3,\ \ 6, 0)$\\
$\lambda_1^1 = (1, 0.2, -4, 0)$\\
\end{tabular}
\end{flushright}
\vspace{-1em}
\caption[DKT Simulation Results - Comparison between true and DKT-estimated biomarker trajectories and subject time-shifts.]{Comparison between true and DKT-estimated subject time-shifts and biomarker trajectories. (top-left/top-middle) Scatter plots of the true shifts (y-axis) against estimated shifts (x-axis), for the 'synthetic AD' and 'synthetic PCA' diseases. We then show the DKT-estimated and true trajectories of the agnostic units within the 'synthetic AD' disease (top-right, "Dis0") and the 'synthetic PCA' disease (bottom-left, "Dis1"). Finally, we also show the biomarker trajectories within unit 0 (bottom-center) and unit 1 (bottom-right). Parameters used for generating the trajectory shapes are shown in the table on the right.}
  \label{fig:dktSynthTrajCompTrue}
\end{figure}

Results on synthetic data in the presence of ground truth (Fig. \ref{fig:dktSynthTrajCompTrue}) suggest that DKT can robustly estimate the trajectory parameters (MAE $<$ 0.058) as well as the subject-specific time-shifts ($R^2$ $>$ 0.98). While some errors in trajectory estimation can be noticed, these are due to the informed priors on the model parameters in order to ensure identifiability and convergence of parameters.

\begin{figure}[htp]
\centering
 \includegraphics[width=0.8\textwidth, trim=0 20 0 0, clip]{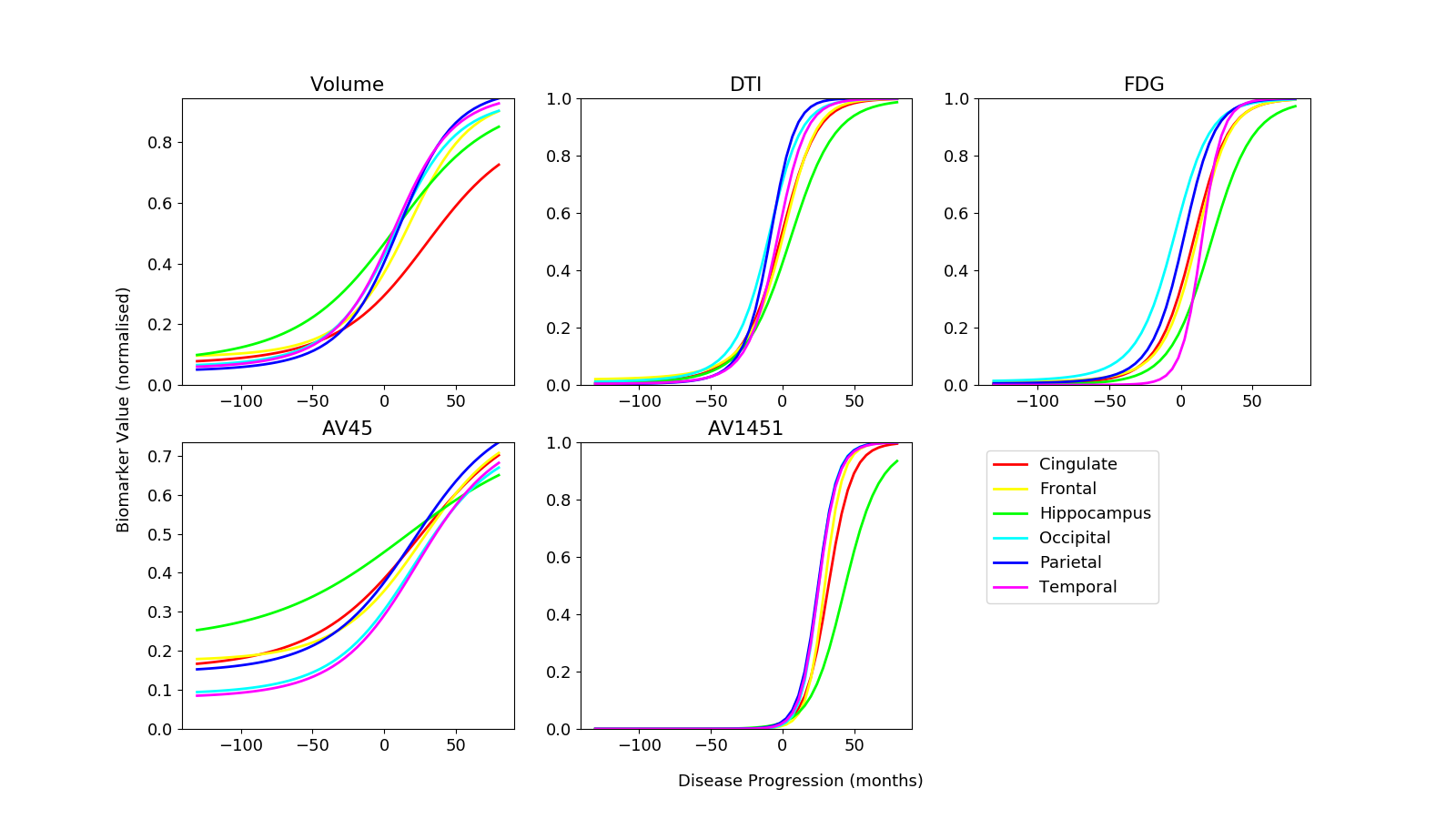}
 \caption{Estimated trajectories for the PCA cohort. The only data that were available were the MRI volumetric data. The dynamics of the other biomarkers has been inferred by the model using data from typical AD, and taking into account the different spatial distribution of pathology in PCA vs tAD. }
 \label{fig:PCAtrajByModality}
\end{figure}

We then apply DKT to real patient data, with the aim of transferring multimodal biomarker trajectories from tAD to PCA. The inferred PCA trajectories, shown in Fig. \ref{fig:PCAtrajByModality}, recapitulate known patterns in PCA \cite{crutch2012posterior}, where posterior regions such as occipital and parietal lobes are predominantly affected in later stages. As opposed to typical AD, we find that the hippocampus is affected later on, further suggesting the model did not transfer too much tAD specific information. Here, we demonstrate the possibility of inferring plausible non-MRI biomarkers in a rare neurodegenerative disease, in lack of such data for these subjects. As far as we are aware, this is the first time a continuous signature of non-MRI biomarkers is estimated for PCA, due to its rarity and lack of data.

\subsection{Validation on DTI Data in tAD and PCA}
\label{sec:dktResVal}

We further validated DKT by predicting unseen DTI data from two patient datasets: 1) TADPOLE subjects with a different progression from the training subjects, and 2) a separate test set of 20 DTI scans from controls and PCA patients from the DRC -- full demographics are given in the supplementary material. To split TADPOLE into subgroups with different progression, we used the SuStaIn model by \cite{young2018uncovering}, which resulted into three subgroups: hippocampal, cortical and subcortical, with prominent early atrophy in the hippocampus, cortical and subcortical regions respectively. To evaluate prediction accuracy, we computed the rank correlation between the DKT-predicted biomarker values and the measured values in the test data. We compute the rank correlation instead of mean squared error as it is not susceptible to systemic biases of the models when predicting "unseen data" in a certain disease.


Validation results are shown in Table \ref{sec:dktPerfMetrics}, for hippocampal to cortical TADPOLE subgroups (other pairs of subgroups not shown due to lack of space) as well as PCA subjects. When predicting missing DTI markers of the TADPOLE cortical subgroup as well as PCA subjects from the DRC cohort (Table \ref{sec:dktPerfMetrics}), the DKT correlations are generally high for the cingulate, hippocampus and parietal, and lower for the frontal lobe. DKT also shows favourable performance compared to four other models: the latent-stage model from \cite{jedynak2012computational}, a multivariate Gaussian Process model with RBF kernel that predicts a DTI ROI marker from multiple MRI markers, as well as cubic spline and linear models that predict a regional DTI biomarker directly from its corresponding MRI marker. In particular for predicting DTI FA in the parietal and temporal lobes, DKT has significantly better predictions that almost all methods tested.

\newcommand{\cw}{c}

\begin{table}
\centering
\fontsize{7}{10}\selectfont
\begin{tabular}{c | c c c c c c}
\textbf{Model} & \textbf{Cingulate} & \textbf{Frontal} & \textbf{Hippocam.} & \textbf{Occipital} & \textbf{Parietal} & \textbf{Temporal}\\
& \multicolumn{6}{c}{\textbf{TADPOLE: Hippocampal subgroup to Cortical subgroup}}\\
DKT (ours) &      0.56 $\pm$ 0.23 &    \textbf{0.35 $\pm$ 0.17} &        \textbf{0.58 $\pm$ 0.14} &     -0.10 $\pm$ 0.29 &     \textbf{0.71 $\pm$ 0.11} &     \textbf{0.34 $\pm$ 0.26} \\
Latent stage &      0.44 $\pm$ 0.25 &    0.34 $\pm$ 0.21 &       0.34 $\pm$ 0.24* &     \textbf{-0.07 $\pm$ 0.22} &     0.64 $\pm$ 0.16 &    0.08 $\pm$ 0.24* \\
Multivariate &      \textbf{0.60 $\pm$ 0.18} &   0.11 $\pm$ 0.22* &       0.12 $\pm$ 0.29* &     -0.22 $\pm$ 0.22 &   -0.44 $\pm$ 0.14* &   -0.32 $\pm$ 0.29* \\
Spline &    -0.24 $\pm$ 0.25* &  -0.06 $\pm$ 0.27* &        0.58 $\pm$ 0.17 &     -0.16 $\pm$ 0.27 &    0.23 $\pm$ 0.25* &    0.10 $\pm$ 0.25* \\
Linear &    -0.24 $\pm$ 0.25* &   0.20 $\pm$ 0.25* &        0.58 $\pm$ 0.17 &     -0.16 $\pm$ 0.27 &    0.23 $\pm$ 0.25* &    0.13 $\pm$ 0.23* \\
& \multicolumn{6}{c}{\textbf{typical Alzheimer's to Posterior Cortical Atrophy}}\\
DKT (ours) &    0.77 $\pm$ 0.11 &    0.39 $\pm$ 0.26 &      0.75 $\pm$ 0.09 &    0.60 $\pm$ 0.14 &    \textbf{0.55 $\pm$ 0.24} &    \textbf{0.35 $\pm$ 0.22} \\
Latent stage &    \textbf{0.80 $\pm$ 0.09} &    \textbf{0.53 $\pm$ 0.17} &      \textbf{0.80 $\pm$ 0.12} &    0.56 $\pm$ 0.18 &    0.50 $\pm$ 0.21 &    0.32 $\pm$ 0.24 \\
Multivariate &   0.73 $\pm$ 0.09 &   0.45 $\pm$ 0.22  &    0.71 $\pm$ 0.08 & -0.28 $\pm$ 0.21* &  0.53 $\pm$ 0.22  &  0.25 $\pm$ 0.23* \\
Spline &   0.52 $\pm$ 0.20* &  -0.03 $\pm$ 0.35* &     0.66 $\pm$ 0.11* &   0.09 $\pm$ 0.25* &    0.53 $\pm$ 0.20 &   0.30 $\pm$ 0.21* \\
Linear &   0.52 $\pm$ 0.20* &    0.34 $\pm$ 0.27 &     0.66 $\pm$ 0.11* &    \textbf{0.64 $\pm$ 0.17} &    0.54 $\pm$ 0.22 &   0.30 $\pm$ 0.21* \\
\end{tabular}
\vspace{0.5em}
\caption[Performance evaluation of DKT and other models]{Performance evaluation of DKT and four other statistical models of decreasing complexity. We show the rank correlation between predicted biomarkers and measured biomarkers in (top) TADPOLE subgroups and (bottom) PCA. (*) Statistically significant difference in the performance of DKT vs the other models, based on a two-tailed t-test, Bonferroni corrected.}
\label{sec:dktPerfMetrics}
\end{table}

\section{Discussion}
\label{sec:dktDis}

In this work we made initial steps at the challenging problem of transfer learning between different neurodegenerative diseases. Our proposed DKT method enabled the estimation of quantitative non-MRI trajectories in a rare disease (PCA) where very limited data was available. To our knowledge, this is the first time a multimodal continuous signature is derived for PCA, as the only other longitudinal study of PCA only computed atrophy measures from MRI scans \cite{lehmann2011cortical}. Our work has however several limitations, which can be addressed in future research: 1) to account for population heterogeneity, DKT can be easily extended to include subject-specific effects; 2) improved schemes for biomarker allocation to agnostic units can take connectivity into account, or derive it from the data automatically; 3) DKT can be further validated on more complex synthetic experiments with a range of datasets generated with different parameters.

\section{Acknowledgements}

This work was supported by the EPSRC Centre For Doctoral Training in Medical Imaging with grant EP/L016478/1 and in part by the Neuroimaging Analysis Center through NIH grant NIH NIBIB NAC P41EB015902. Data collection and sharing for this project was funded by the Alzheimer’s Disease Neuroimaging Initiative (ADNI) (National Institutes of Health Grant U01 AG024904) and DOD ADNI (Department of Defense award number W81XWH-12-2-0012). The Dementia Research Centre is an ARUK coordination center.

\bibliographystyle{unsrtnat}

\clearpage

\section{Supplementary material}

\subsection{Parameter Estimation}

\newcommand{\uu}{^{(u)}}
\newcommand{\um}{^{(u-1)}}

We estimate the model parameters using a two-stage approach. In the first stage, we perform belief propagation within each agnostic unit and then within each disease model. In the second stage we jointly optimise across all agnostic units and disease models using loopy belief propagation. An overview of the algorithm is given in Figure \ref{fig:dktAlgo}. Given the initial parameters estimated from the first stage (line 1), the algorithm continuously updates the biomarker trajectories within the agnostic units (lines 4-5), dysfunction trajectories (line 8) and subject-specific time shifts (line 10) until convergence. The cost function for all parameters is nearly identical, the main difference being the measurements $(i,j,k)$ over subjects $i$, visits $j$ and biomarkers $k$ that are selected for computing the measurement error. For estimating the trajectory of biomarker $k$ within agnostic unit $\psi(k)$, measurements are taken from $\Omega_k$ representing all measurements of biomarker $k$ from all subjects and visits. For estimating the dysfunction trajectories, $\Omega_{d,l}$ represents the measurement indices from all subjects with disease $d$ (i.e. $d_i = d$) and all biomarkers $k$ that belong to agnostic unit $l$ (i.e. $\psi(k) = l$). Finally, $\Omega_i$ (line 10) represents all measurements from subject $i$, for all biomarkers and visits.

\begin{figure}
\begin{algorithm}[H]
\scriptsize
 Initialise $\boldsymbol{\theta}^{(0)}$, $\boldsymbol{\lambda}^{(0)}$, $\boldsymbol{\beta}^{(0)}$\\
  \While{$\boldsymbol{\theta}$, $\boldsymbol{\lambda}$, $\boldsymbol{\beta}$ not converged}{
   \tcp*[l]{Estimate biomarker trajectories (disease agnostic)}
    \For{$k=1$ to $K$}{
      ${\theta_k\uu = \argmin_{\theta_k} \sum_{(i,j) \in \Omega_k} \left[y_{ijk} - g\left(f(\beta_i\um + m_{ij}; \lpum) ; \theta_k\right) \right]^2  - log\ p(\theta_k)}$\\
      ${\epsilon_k\uu = \frac{1}{|\Omega_k|} \sum_{(i,j) \in \Omega_k}    \left[y_{ijk} - g\left(f(\beta_i\um + m_{ij}; \lpum) ; \theta_k\uu \right) \right]^2 }$\\
    }
     \tcp*[l]{Estimate dysfunction trajectories (disease specific)} 
    \For{$d=1 \in \mathbb{D}$}{
      \For{$l=1 \in \Lambda$}{
        ${\lambda_{d}^{l, (u)} = \argmin_{\lambda_{d}^{l}} \sum_{(i,j,k) \in \Omega_{d,l}} \left[y_{ijk} - g\left(f(\beta_i\um + m_{ij}; \lambda_{d}^{l}) ; \theta_k\uu 
        \right) \right]^2  - log\ p(\lambda_{d}^{l})}$\\
      }
    }
    \tcp*[l]{Estimate subject-specific time shifts} 
    \For{$i=1 \in [1, \dots, S]$}{
      ${\beta_i\uu = \argmin_{\beta_i} \sum_{(j,k) \in \Omega_i} \left[y_{ijk} - g\left(f(\beta_i + m_{ij}; \lpuu) ; \theta_k\uu
      \right) \right]^2  - log\ p(\beta_i)}$\\
    }
}
\normalfont
\end{algorithm}
\caption[The algorithm for estimating the DKT parameters]{The algorithm used to estimate the DKT parameters, based on loopy belief-propagation.}
\label{fig:dktAlgo}
\end{figure}

\subsection{Generation of synthetic dataset}

We tested DKT on synthetic data, to assess its performance against known
ground truth. More precisely, we generated data that follows the DKT model
exactly, and tested DKT's ability to recover biomarker trajectories and subject time-shifts. 

We generated the synthetic data as follows, using parameters from Table \ref{tab:synParams}:
\begin{itemize}
 \item We simulate two synthetic diseases, "synthetic PCA" and "synthetic AD"
 \item We define 6 biomarkers that we allocate to agnostic units $l_0$ and $l_1$ (Table \ref{tab:synParams} top)
 \item Within each agnostic unit, we define the parameters $\{\theta_0$, ..., $\theta_5\}$ corresponding to biomarker trajectories within the agnostic unit.
 \item For each disease, we define the parameters $\lambda$ corresponding to trajectories of dysfunction scores.
 \item We then sample data from 100 synthetic AD subjects and 50 PCA subjects with $\beta_i$ as given in Table \ref{tab:synParams} bottom using the model likelihood (Eq. 2 from main paper). For each subject, we generate data for 4 visits, each 1 year apart.
\end{itemize}

 \begin{table}
 \centering
 \begin{tabular}{c | c}
& \textbf{Trajectory parameters} \\
 Biomarker allocation &  $l_0:\{k_0, k_2, k_4\}$, $l_1: \{k_1, k_3, k_5\}$ \\
Agnostic unit $l_0$ &  $\theta_0 = (1,5,0.2,0)$, $\theta_2 = (1,5,0.55,0)$,  $\theta_4 = (1,5,0.9,0)$  \\
Agnostic unit $l_1$ & $\theta_1 = (1,10,0.2,0)$, $\theta_3 = (1,10,0.55,0)$, $\theta_5 = (1,10,0.9,0)$ \\
  
"Synthetic AD" & $\lambda_0^0 = (1, 0.3, -4, 0)$  and $\lambda_0^1 = (1, 0.2, 6, 0)$ \\
 "Synthetic PCA" & $\lambda_1^0 = (1, 0.3, 6, 0)$ and $\lambda_1^1 = (1, 0.2, -4, 0)$ \\
\hline
& \textbf{Subject parameters} \\
 Number of subjects & 100 (synthetic AD) and 50 (synthetic PCA) \\ 
 Time-shifts $\beta_i$ & $\beta_i \sim U(-13,10)$ years \\
 Diagnosis & $p(control) \propto Exp (-4.5)$,  $p(patient) \propto Exp (4.5)$\\
 Data generation & 4 visits/subject, 1 year apart, $\epsilon_k = 0.05$\\ 
\end{tabular}
\caption{Parameters used for synthetic data generation, emulating the TADPOLE and DRC datasets.}
\label{tab:synParams}
\end{table}

\subsection{Demographics of test sets}

The cohort from the Dementia Research Centre UK used for validation consisted of 10 subjects diagnosed with Posterior Cortical Atrophy, with a mean age of 59.4, 40\% females, as well as 10 age-matched controls with a mean age of 59.3, 50\% females.

For the validation on TADPOLE subgroups, we used applied the SuStaIn model on TADPOLE to split the population into three subgroups with different progression: hippocampal, cortical and subcortical subypes with prominent atrophy in the hippocampus, cortical and subcortical areas respectively.
The resulting subgroups had the following demographics:

\begin{table}
 \begin{tabular}{c | c c c c}
\textbf{Cohort} & \textbf{Nr. subjects} & \textbf{Nr. visits} & \textbf{Age (baseline)} & \textbf{Gender (\%F)}\\
\hline
Controls (Hippocampal) & 31 & 2.3 $\pm$ 1.8 & 74.4 $\pm$ 6.9 & 38\%\\
AD (Hippocampal) & 21 & 1.5 $\pm$ 0.8 & 74.5 $\pm$ 5.5 & 42\%\\
\hline
Controls (cortical) & 21 & 2.3 $\pm$ 1.3 & 70.9 $\pm$ 5.4 & 42\%\\
AD (cortical) & 35 & 1.7 $\pm$ 0.9 & 72.8 $\pm$ 7.4 & 28\%\\
\hline
Controls (subcortical) & 28 & 3.0 $\pm$ 1.5 & 73.7 $\pm$ 6.5 & 42\%\\
AD (subcortical) & 27 & 1.6 $\pm$ 0.9 & 73.7 $\pm$ 7.5 & 33\%\\
 \end{tabular}
 \caption{Demographics of the subjects in the three TADPOLE subgroups.}
 \label{tab:demogTadSubtypes}
\end{table}

\end{document}